\documentclass[11pt,a4paper]{article}
\usepackage[hyperref]{emnlp2020}
\usepackage{times}
\usepackage{latexsym}

\usepackage{graphicx}
\usepackage{subfigure}
\usepackage{amsmath}
\usepackage{amsfonts}
\usepackage{multirow}
\usepackage{CJKutf8}

\usepackage{color}
\usepackage{listings}
\usepackage{xcolor}

\usepackage{microtype}

\aclfinalcopy % Uncomment this line for the final submission
\title{Long-Short Term Masking Transformer: A Simple but Effective Baseline\\for Document-level Neural Machine Translation}

\author{
Pei Zhang, Boxing Chen, Niyu Ge, Kai Fan\thanks{~~corresponding author.}\\
Alibaba Group Inc.\\
\texttt{\{xiaoyi.zp,boxing.cbx,niyu.ge,k.fan\}@alibaba-inc.com} 
}

\date{}

\begin{document}
\maketitle
\begin{abstract}
Many document-level neural machine translation (NMT) systems have explored the utility of context-aware architecture, usually requiring an increasing number of parameters and computational complexity. 
However, few attention is paid to the baseline model. 
In this paper, we research extensively the pros and cons of the standard transformer in document-level translation, and find that the auto-regressive property can simultaneously bring both the advantage of the consistency and the disadvantage of error accumulation. 
Therefore, we propose a surprisingly simple long-short term masking self-attention on top of the standard transformer to both effectively capture the long-range dependence and reduce the propagation of errors. 
We examine our approach on the two publicly available document-level datasets. 
We can achieve a strong result in BLEU and capture discourse phenomena.
\end{abstract}

\section{Introduction}

Recent advances in deep learning have led to significant improvement of Neural Machine Translation (NMT) \cite{sutskever2014sequence,bahdanau2014neural,luong2015effective,vaswani2017attention}. 
Particularly, the performance on the sentence-level translation of both low- and high- resource language pairs is dramatically improved \cite{kudugunta2019investigating,lample2018phrase,lample2019cross}. 
However, when translating text with long-range dependencies, such as in conversations or documents, the original mode of translating one sentence at a time ignores the discourse phenomena \cite{voita2019context,voita2019good}, introducing undesirable behaviors such as inconsistent pronouns across different translated sentences. 

Document-level NMT, as a more realistic translation task in these scenarios, has been systematically investigated in the machine translation community. 
Most literatures focused on looking back a fixed number of previous source or target sentences as the document-level context \cite{tu2018learning,voita2018context,zhang2018improving,miculicich2018document,voita2019context,voita2019good}. 
Some latest works innovatively attempted to either get the most out of the entire document context or dynamically select the suitable context \cite{maruf2018document,yang2019enhancing,maruf2019selective,jiang2019document}. 
Because of the scarcity of document training data, the benefit gained from such an approach, as reflected in BLEU, is usually limited.
We therefore elect to pay attention to the context in the previous $n$ sentences only where $n$ is a small number and usually does not cover the entire document.
  
Almost all of the latest studies chose the standard transformer model as their baseline which translates each sentence in the document with the model trained on the sentence-level data.  
The cohesion and consistency are in general poor.  
A more reasonable baseline is to train the transformer with the context prepended, and this modification could be simply implemented via data preprocessing. 
\citet{bawden2018evaluating} conducted a detailed analysis of RNN-based NMT models on the topic of whether or not to include the extended context. 
Consistency and precision is often viewed as a trade-off of each other.
We conduct a detailed analysis of the effect of document context on consistency in transformer architecture accepting multi-sentence input.

When it comes to leveraging the contextual information, the common approach is to model the interaction between the sentence and its context with specially designed attention modules \cite{kim2019and}. 
Such works tend to include more than one encoder or decoder, with a substantial number of parameters and additional computations. 
In our work, we reduce the contextual and regular attention modules into one single encoder and decoder. 
Our idea is motivated by the one transformer decoder with the two-stream self-attention \cite{yang2019xlnet}. 
%\cite{zhang2019lattice,fan2019neural}
In particular, we maintain two different sets of hidden states and employ two different masking matrices to capture the long and short term dependencies. 

The contributions of this paper are threefold:  
i) we extensively research the performance of the standard transformer in the setting of multi-sentence input and output; 
ii) we propose a simple but effective modification to adapting the transformer for document NMT with the aim of ameliorating the effect of error accumulation; 
iii) our experiments demonstrate that even the simple baseline can achieve comparable results.

\section{The Proposed Approach}

The standard transformer NMT follows the typical encoder-decoder architecture with using stacked self-attention, pointwise fully connected layers, and the encoder-decoder attention layers. 
The self-attention in the decoder allows only those positions from the left up to the current one to be attended to, preventing information flow to the right beyond the current target and preserving the auto-regressive property. 
The illegal connections will be masked out by setting as $-\infty$ before the softmax operation.
The attention probability can be succinctly written in a unified formulation.
\vspace{-2mm}
\begin{equation}
\vspace{-2mm}
A = \textbf{Softmax}\left(\frac{Q K^\top}{\sqrt{d/h}} + M \right) \label{eq:attn} 
\end{equation}
where the matrices $Q, K$ represent queries and keys in attention module \cite{vaswani2017attention}, and $M$ is the masking matrix. 
For the encoder self-attention and the encoder-decoder attention, $M = \mathbf{0}$. 
For the decoder self-attention, $M$ is an upper triangular matrix with zero on the diagonal and non-zero ($-\infty\approx-10^9$) everywhere else.

\subsection{Long-Short Term Masking Transformer}

\begin{figure}[t]
\vspace{-3mm}
\includegraphics[width=1.1\columnwidth]{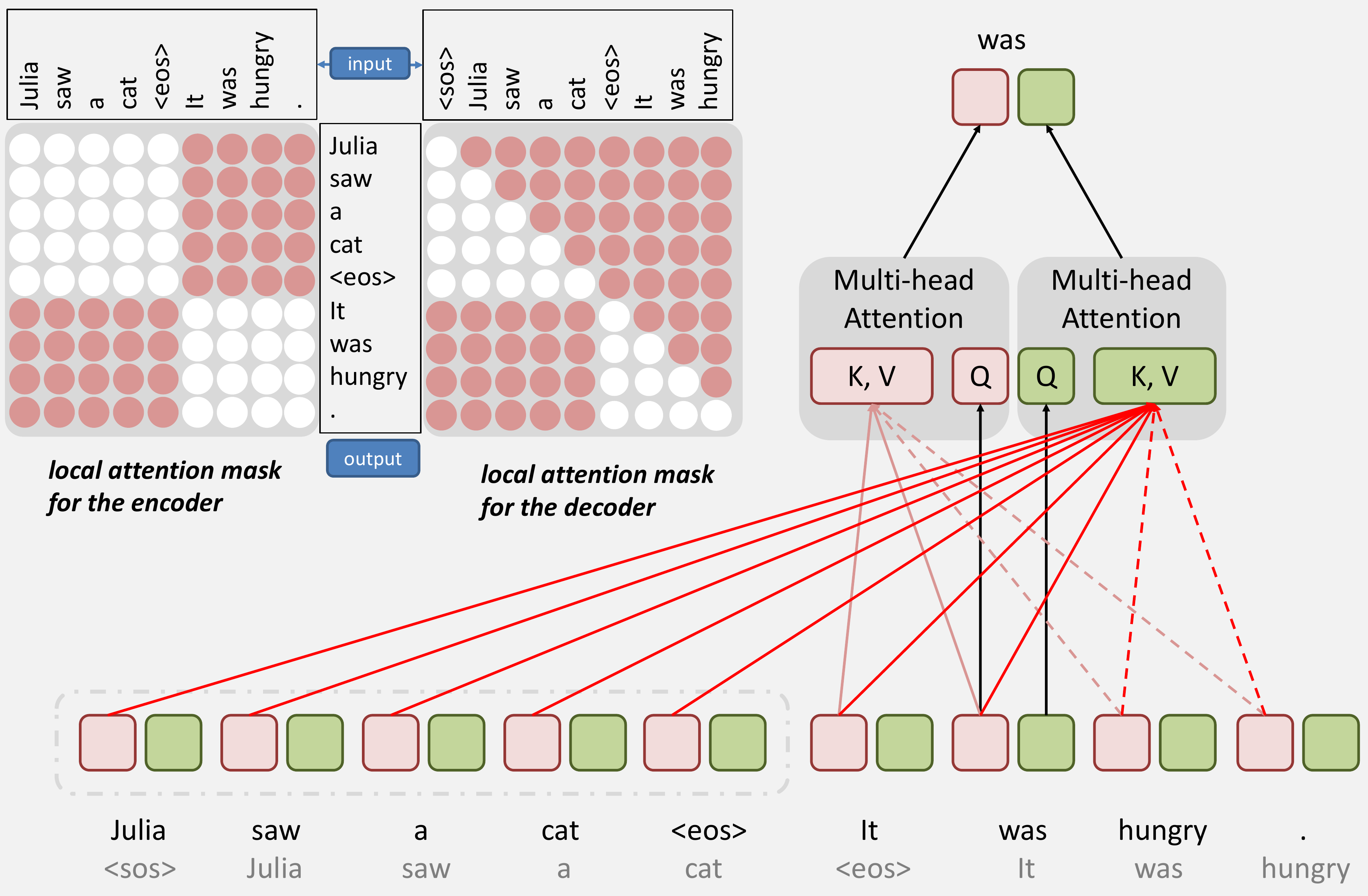}
\caption{Illustration of the Long-Short Term Masking Self-Attention. Green nodes: global self-attention, which is the same as the standard self-attention. Pink nodes: local self-attention, which does not have access to the information from the document context. The red dash lines is removed in the decoder attention.}
\label{fig:twostream}
\end{figure}

The basic setup in this work is multi-sentence input and output, denoted as $k$-to-$k$ model. 
In other words, both the encoder and decoder need to consume $k$ sentences during training and inference. 
Therefore, in our modified transformer, the regular self-attention is substituted by the long-short term masking self-attention (illustrated in Figure~\ref{fig:twostream}). 
While the idea of most context-aware model is to introduce another isolated attention module, we propose to maintain two stream attentions via the local and global representations, but the parameters to calculate queries, keys and values are shared.

The global self-attention, simply following the calculation in Eq~(\ref{eq:attn}), serves a similar role to the standard hidden states in transformer. 
The keys and values can broadly look around from the first token to the last one, and the global hidden state of the next layer will summarize the information of both the context and current sentence. 
The query vector directly comes from the global hidden states of the previous layer via a fully connect layer.

The local self-attention only accesses the information of the current sentence, where the contextual information from the previous sentence(s) is blocked when computing the keys and values. 
Similar to the masking strategy of the transformer decoder, the implementation of the local self-attention is to mask out the tokens of the context via $-\infty$ inside the scaled dot-product operation. 
Figure~\ref{fig:twostream} depicts the masking matrices of the local self-attention for the encoder and decoder respectively. 
They are both diagonal block matrices, where each block represents the local self-attention of current sentence and the blank and maroon dots denote value 0 and $-\infty$. 
When calculating attention weights, we only need to replace the $M$ in Eq~(\ref{eq:attn}) with the block masking matrices.

For the two sets of hidden representations in the final layer, we can either aggregate them with element-wise operation (such as summation or concatenation) or directly use global states to predict the distribution of target language model. 
In our work, we adopt the concatenation, and subsequently transform them via a fully connected layer to reduce dimensionality.

\begin{figure*}[t]
\vspace{-7mm}
\centering
\subfigure[Standard Transformer]{
\includegraphics[height=47mm]{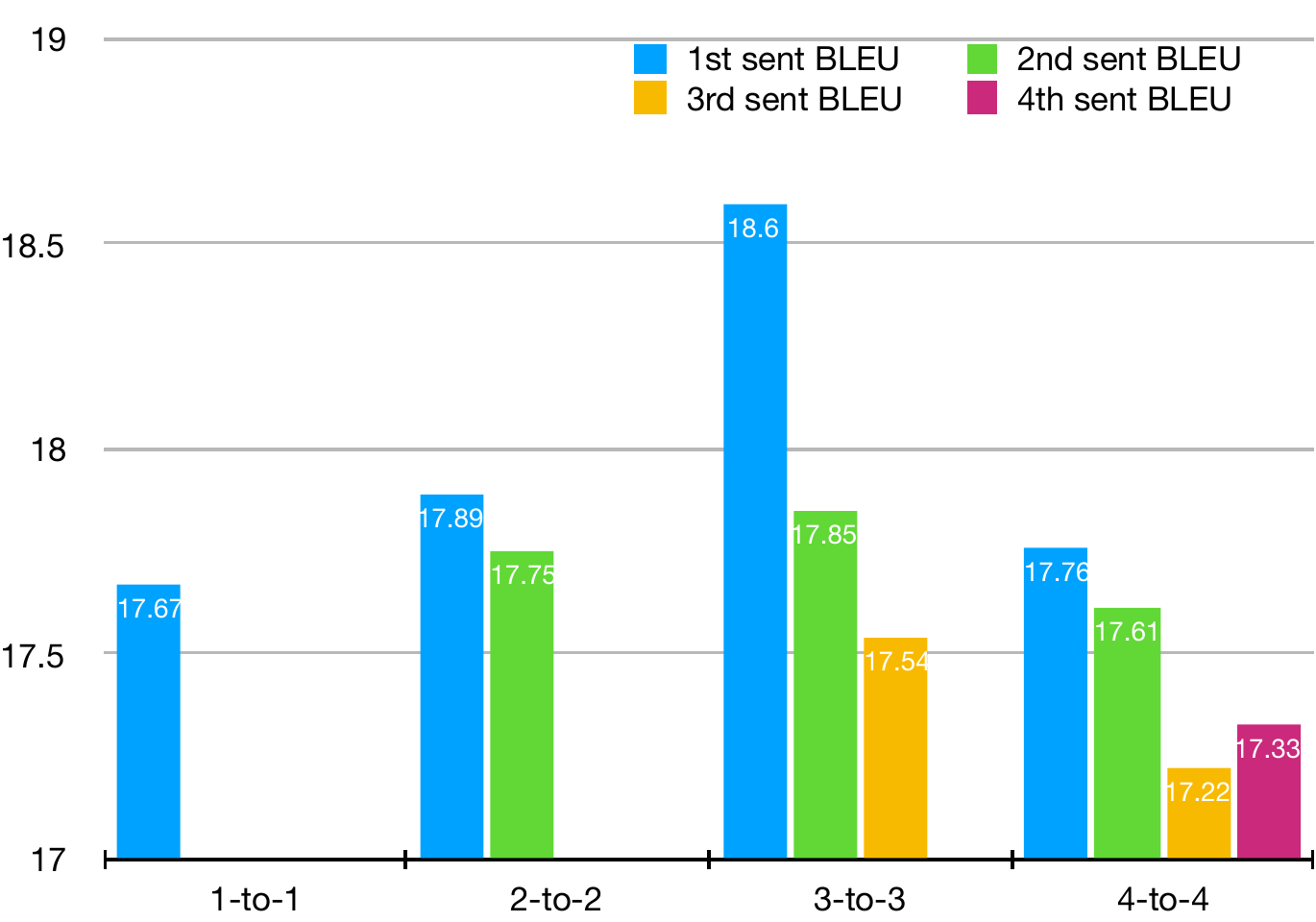}
}
\subfigure[Our Approach]{
\includegraphics[height=47mm]{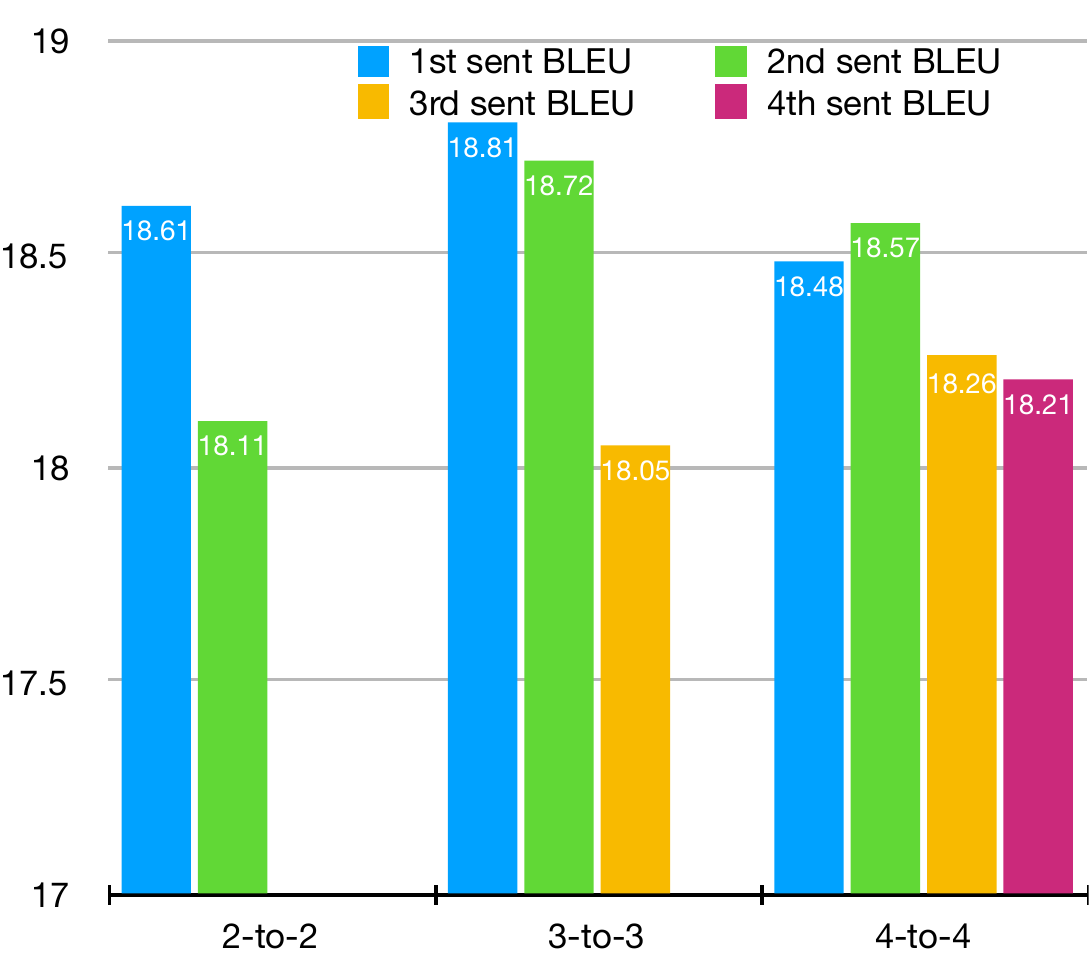}
}
\subfigure[Other works]{
\includegraphics[height=47mm]{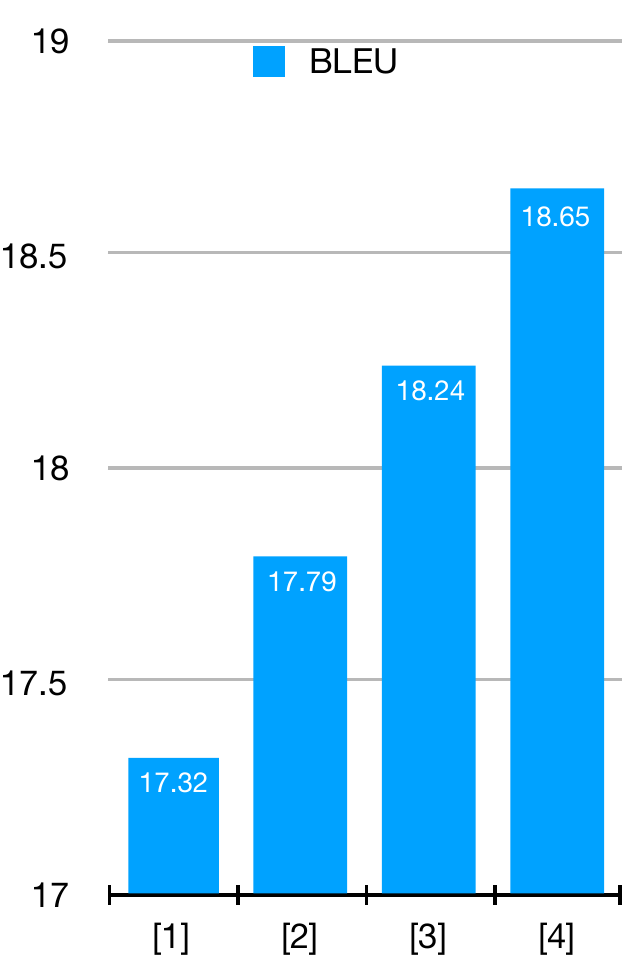}
}
\caption{Zh-En: The $j$-th sentence BLEU of $k$-2-$k$ model, where it means the average BLEU on the $j$-th sentence.  [1] \cite{tu2018learning} [2] \cite{miculicich2018document} [3] \cite{voita2018context} [4] \cite{jiang2019document}}
\label{fig:bleu}
\end{figure*}

\begin{table}[t]
\vspace{-5mm}
\scriptsize
\centering
%\resizebox{\columnwidth}{!}{
\begin{tabular}{c|l}
\hline
\multirow{2}{*}{Src} & \begin{CJK}{UTF8}{gbsn}“在死之前，我想种一棵树” “在死之前，我想过隐居生活”\end{CJK} \\
 & \begin{CJK}{UTF8}{gbsn}“在死之前，我想在抱她一次”\end{CJK} \\
\hline
\multirow{2}{*}{Ref} & ``\textcolor{orange}{Before I die, I want to} plant a tree." ``\textcolor{orange}{Before I die, I want to} live \\
 &  off the grid." ``\textcolor{orange}{Before I die, I want to} hold her one more time." \\
\hline
\multirow{2}{*}{Sys0} & ``Before death, I want a tree." ``Before I die, I want to live lives." \\
 & ``Before death, I want to hug her again." \\
\hline
\multirow{2}{*}{Sys1} & I want to be a tree before I die. ``Before death, I want to become \\
 & invisible." ``Before death, I want to hug her again." \\
\hline
\multirow{2}{*}{\textbf{Sys2}} & ``I want to create a tree before I die." ``Before I die, I want to \\
 & live a hidden life." ``Before I die, I want to hug her again." \\
\hline
\hline
\multirow{2}{*}{Src} & \begin{CJK}{UTF8}{gbsn}在左边你能看到一个小船。这是一个约15英尺的船。我想让\end{CJK} \\
 & \begin{CJK}{UTF8}{gbsn}你们注意冰山的形状它在水面上的变形。\end{CJK} \\
\hline
\multirow{3}{*}{Ref} & You can see on the left side a small \textcolor{orange}{boat}. That's about a 15 foot   \\
 & \textcolor{orange}{boat}. And I'd like you to pay attention to the shape of the iceberg \\
 & and where it is at the waterline. \\
\hline
\multirow{2}{*}{Sys0} & On the left you see a small boat. It's about 15 feet. I want you to  \\
 & look at the shape of the iceberg that it deformed on the water. \\
\hline
\multirow{3}{*}{Sys1} & On the left you can see a small boat. This is a ship about 15 feet. \\
 & I want you to notice the shape of the iceberg which is distorted \\
 & on the water. \\
\hline
\multirow{3}{*}{\textbf{Sys2}} & On the left you see a small boat. This is a 15 foot boat. I want \\
 & you to pay attention to the shape of the iceberg that's distorted \\
 & on the surface of the water. \\
\hline
\end{tabular}
%}
\caption{Examples of translation results. Sys0: 1-to-1 transformer. Sys1: 3-to-3 transformer. Sys2: 3-to-3 long-short term masking transformer.}
\label{tab:examples}
\end{table}

\section{Experiments}

\textbf{Experimental Setup} 

We carry out experiments with the Chinese-English IWSLT TED talks dataset\footnote{\href{https://wit3.fbk.eu}{https://wit3.fbk.eu}} and English-Russian open-subtitle dataset\footnote{\href{https://github.com/lena-voita/good-translation-wrong-in-context}{https://github.com/lena-voita/good-translation-wrong-in-context}}. 
The widely used Zh-En IWSLT dataset contains around 200K training sentence pairs divided into 1713 documents. 
As is the convention, dev2010 and tst2010-2013 are used for validation and testing respectively. 
The En-Ru subtitle dataset contains around 1.5M conversations, where each conversation includes exactly 4 sentences. 
Two randomly selected subsets of 10,000 instances from movies not included in the training are used for development and test\footnote{\href{http://data.statmt.org/acl18_contextnmt_data/}{http://data.statmt.org/acl18\_contextnmt\_data/}}. 

The BPE tokenization is separately learnt with 32K operations for each language in the dataset. 
The resulting source / target vocabulary sizes for En-Zh and En-Ru datasets are 10296 / 16018 and 12273 / 22642, respectively. 
The token-level batch sizes are 8192 and 16384 for training the Zh-En and En-Ru datasets on two and four P-100 GPUs. 

The model hyper-parameters and the optimizer of standard transformer baseline follow the base setting in \cite{vaswani2017attention}. 
We set the layers in encoder and decoder to 6, and the attention heads to 8. 
The dimensionality of input and output is 512. 
In addition, we add a feed-forward layer before the decoder output layer, with dimensionality 1024, to combine the local and global stream. 
We use the Adam optimizer with $\beta 1$ = 0.9, $\beta 2$ = 0.98 and $\epsilon$ = $10^{-9}$, with 16000 warm-up steps and scale of 4. The batch size for each GPU is 4000.

BLEU score is calculated with the script \texttt{mteval-v13a.pl} in Moses\footnote{\href{https://github.com/moses-smt/mosesdecoder/blob/master/scripts/generic/mteval-v13a.pl}{https://github.com/moses-smt/mosesdecoder/blob/master/scripts/generic/mteval-v13a.pl}}. 
All reported values are evaluated on the test set with the best checkpoint on the development set.

\subsection{Evaluation on BLEU}

We first conduct a detailed analysis on the $k$-to-$k$ translation model with respect to the IWSLT Zh-En dataset. 
In this scenario, the $k$ source and target sentences are concatenated as the input and output to train the transformer. 
During inference, for every consecutive $k$ source sentences, the model produces $k$ target sentences. To translate a test set in a $k-to-k$ model, we keep a sliding window of size $k$. Each sentence is translated $k$ times (excpet for the first $k-1$ sentences), each time as a $j^{th}$ ($j \leq k$) sentence. For example, in a 4-to-4 model, sentence 5 is translated 4 times -- the $1^{st}$ time as the last sentence in the chunk $s_{2},s_{3},s_{4}, s_{5}$, the $2^{nd}$ time as the $3^{rd}$ sentence in the chunk $s_{3},s_{4},s_{5}, s_{6}$, and so on. We thus can assemble different versions of the final translated test set where each sentence is translated as the $j^{th}$ sentence ($j \leq k$) in the translation process. Each of these final documents is evaluated separately. The results are illustrated in Figure~\ref{fig:bleu}. 

\begin{table*}[t]
\vspace{-3mm}
\small
\setlength{\tabcolsep}{2pt}
\centering
%\resizebox{\columnwidth}{!}{
\begin{tabular}{l|c|c|c|c|c|c|c|c}
\hline
\multirow{2}{*}{Models} & Model & \multirow{2}{*}{Beam} & \multirow{2}{*}{multi-bleu} & \multirow{2}{*}{mteval-v13a} & \multirow{2}{*}{\textbf{Deixis}} & \textbf{Lexical} & \textbf{Ellipsis} & \textbf{Ellipsis} \\
& Size & & & & & \textbf{Cohesion} & \textbf{(VP)} & \textbf{(Infl.)} \\
\hline
s-hier-to-2.tied \cite{bawden2018evaluating} & NA & 4 & 26.68 & NA & 60.9 \% &  48.9\% & 65.6\% & 66.4\% \\
\hline
Sentence baseline \cite{voita2019good} & 256M & 4 & 32.40 & NA & 50.0\% & 45.9\% & 28.9\%  & 53.0\% \\
Concat Baseline \cite{voita2019good} & 256M & 4 & 31.56 & NA & 83.5\% & 47.5\% & 76.2\% & 76.6\% \\
CADec \cite{voita2019good} & 458M & 4 & 32.38 & NA & 81.6\% & 58.1\% & 80.0\% & 72.2\% \\
\hline
Concat Baseline \cite{jean2019fill} & 256M & 8 & NA & 31.00 & 83.4\% & 48.9\% & 73.8\% & 76.0\% \\
Partial Copy \cite{jean2019fill} & 256M & 8 & NA & 31.60 & 86.6\% & \textbf{74.9}\% & 77.9\% & 75.5\% \\
\hline
%Our Concat Baseline & 256M & 4 & & & & & & \\
%\hline
\multirow{2}{*}{Our Approach (4-to-4)} & \multirow{2}{*}{262M} & 4 & 31.84 & 32.60 & \textbf{91.0}\% & 46.9\% & 78.2\% & \textbf{82.2}\% \\
                     & & 8 & 32.02 & \textbf{32.80} & & & & \\
\hline
Our Approach (4-to-4) & \multirow{2}{*}{262M} & 4 & 31.31 & 32.28 & 90.5\% & 73.9\% & \textbf{81.0}\% & 80.6\% \\
+ Partial Copy        & & 8 & 31.60 & 32.17 & & & & \\
\hline
\end{tabular}
%}
\caption{En-Ru: The comparison on the accuracy of four consistency metrics. \textbf{i)} multi-bleu are as reported in the original paper. We opt for mteval-v13a because it does not depend on tokenization. \textbf{ii)} Beam size won't affect the values of consistency metrics. \textbf{iii)} Concat Baseline means standard transformer with 4-to-4. }
\label{tab:consistency}
\end{table*}

We can make two inferences from the results.
First, with the Standard transformer, the $1^{st}$ sentence BLEU always the highest (Figure~\ref{fig:bleu}(a)).
This is likely the results of error propagation to subsequent sentences from the auto-regressive property mentioned above.
Second, larger $k$, i.e. more contextual information will not necessarily result in better BLEU score. In this case, $k=2$ or $3$ is better than $k=4$.
We hypothesize that training with longer sentences requiring learning longer range dependencies is fundamentally difficult, especially for such a small dataset. 

When we compare the results of our model with the standard transformer, we have two other findings. 
First, the BLEU scores of our $k$-to-$k$ model outperform those of the standard transformer, and for the $j$-th sentence BLEU, the score does not decline as much as in the standard transformer. 
We believe that our long-short term masking self-attention can, to some extent, relieve the effect of error accumulation. 
Second, when document information is used (i.e., $k>1$), decoding each sentence as the last sentence (ie. using all previous context) achieves higher BLEU scores than decoding each sentence individually in the  standard transformer. 
We pay more attention to the last sentence because presumably it has the richest contextual information; this is also the setting for the results in the next section.

Two qualitative examples are shown in Table~\ref{tab:examples} (more examples can see in the supplementary materials). 
In the first case, compared to Sys0 and Sys1, Sys2 is more consistent in the segments ``Before I die" and ``I want to" of three sentences.
In the second case, the translation of ``boat'' in Sys1 or Sys0 is either omitted or inconsistent in the second sentence, while Sys2 performs better in consistency. 

\subsection{Evaluation on Consistency}

The publicly available open-subtitle En-Ru dataset has a special test data to evaluate consistency of document-level translation systems. The details of the data can be found in \citet{voita2019good}. 
The context of the training and test data contains exactly 3 sentences, so we mainly adopt a 4-to-4 model in our experiments and each sentence is translated as the last sentence in a chunk of 4 sentences.
In this section, we follow previous works to focus on the accuracy of Deixis, Lexical cohesion, Verb phrase ellipsis and Ellipsis (inflection) \footnote{See a short introduction in the supplementary materials.}. 

In Table~\ref{tab:consistency}, we summarize the results of BLEU as well as consistency performance.
s-hier-to-2.tied \cite{bawden2018evaluating} is an RNN-based NMT, so its performance is relatively worse than the other transformer-based models. 
In contrast, our approach can achieve better performance with respect to both BLEU and consistency, except for lexical cohesion. 
Especially the accuracy of lexical cohesion of Partial Copy \cite{jean2019fill} exceeds ours by a large margin. 
\citet{jean2019fill} filled the missing context with partial copy strategy, since the repetition can naturally enhance the lexical cohesion. 
Therefore, when we also apply the partial copy trick to our model, the lexical cohesion can boost by 27\% but the BLEU is sacrificed. 
The Lexical Cohesion of CADec \cite{voita2019good} is a bit higher than our approach without partial copy. 
Considering that CADec is almost double-sized of our standard transformer and complicated architecture with the backbone of the deliberation networks \cite{xia2017deliberation}, the gain over baseline is much higher cost than ours. 
In summary, our model can achieve a strong result in both BLEU and consistency with few extra model parameters.

\section{Discussions and Conclusions}

In this work, we present a simple but effective variation with the long-short term masking strategy, and we performed comparative studies with the $k$-to-$k$ translation model of the standard transformer.
Just as the big, complex neural network architectures with great many parameters has its power, small but efficient modification like ours to the classical transformer has its unique appeals. 
Other examples of simple but impactful ideas are data augmentation and the round-trip back-translation \cite{voita2019context}, to name just a few. 
Big or small, complex or simple, each has its distinct advantages. 
We're encouraged by our findings that in tandem with the great machinery that could bring powerful results, simplistic approaches could be just as efficacious.

\section*{Acknowledgments}

This work is partly supported by National Key R\&D Program of China (2018YFB1403202).

\bibliographystyle{acl_natbib}
\bibliography{emnlp2020}

\appendix

\section{Evaluation Metrics of Consistency}

BLEU is a commonly used metric to evaluate the precision-based quality of the translation in terms of $n$-gram, but it is not fit to evaluate discourse phenomena, because $n$-gram precision does not specifically reflect the cohesion and consistency in the long-range dependencies. 
\textbf{Deixis} addresses the error related to personal pronouns, specifically gender marks and informal/formal distinction. 
\textbf{Lexical cohesion} is refers to the consistency of a word or phrase when it occurs multiple times.
\textbf{Ellipsis} is the omission of words that are understood from the context and it sometimes involves replacement of generic term for a specific term (such as 'did' for 'saw' in English). Since the target language is Russian, we care about both the verb and inflection.

\section{Code in TensorFlow}

We present the code snippet for generating local masking matrix for transformer encoder. 
The matrix for transformer decoder is simply add the above encoder matrix to the regular decoder self-attention masking matrix.

\lstdefinestyle{lfonts}{
basicstyle = \scriptsize\ttfamily, 
stringstyle = \color{purple}, 
keywordstyle = \color{blue!60!black}\bfseries, 
commentstyle = \color{olive}\scshape, 
} 
\lstdefinestyle{lnumbers}{
numbers = left, 
numberstyle = \tiny, 
numbersep = 1em, 
firstnumber = 1,
stepnumber = 1, 
} 
\lstdefinestyle{llayout}{
breaklines = true, 
tabsize = 2, 
columns = flexible, 
} 
\lstdefinestyle{lgeometry}{ 
xleftmargin = 20pt, 
xrightmargin = 0pt, 
frame = tb, 
framesep = \fboxsep, 
framexleftmargin = 20pt, 
} 
\lstdefinestyle{lgeneral}{ 
style = lfonts,
style = lnumbers, 
style = llayout,
style = lgeometry, 
} 
\lstdefinestyle{python}{ 
language = {Python}, 
style = lgeneral, 
}

\begin{lstlisting}[style = python]
def generate_masking(inputs, sentence_sep_id):
    """Generate Long Short Term Masking
    Args:
        inputs: a dense vector [batch, length] of source or target word ids
        sentence_sep_id: the id of the sentence separation token
    """
    shape = tf.shape(inputs)
    length = shape[1]
    sentence_sep_id_matrix = sentence_sep_id * tf.ones(shape, dtype=inputs.dtype)
    sentence_end = tf.cast(tf.equal(inputs, sentence_sep_id), tf.float32)
    sentence_end_mask = tf.cumsum(sentence_end, axis = -1)
    sentence_end_mask_expand_row = tf.expand_dims(sentence_end_mask, -1)
    sentence_end_mask_expand_row = tf.tile(sentence_end_mask_expand_row, [1, 1,  length])
    sentence_end_mask_expand_column = tf.expand_dims(sentence_end_mask, -2)
    sentence_end_mask_expand_column = tf.tile(sentence_end_mask_expand_column, [1, length, 1])
    mask = tf.cast(tf.equal(sentence_end_mask_expand_row, sentence_end_mask_expand_column), tf.float32)
    mask = -1e9 * (1.0 - mask)
    mask = tf.reshape(mask, [-1, 1, length, length])

    return mask
\end{lstlisting}

\section{More Examples}

We randomly selected three translation examples and illustrated in Table~\ref{tab:examples}. 
For Example 1, the proposed system learnt ``And" at the beginning of the translation, which is a side effect of document-level training. 
For Example 2, whether using ``love" or ``love to" is consistency in the proposed system and 1-to-1 baseline transformer. It seems that 1-to-1 baseline can approximately translate \begin{CJK}{UTF8}{gbsn}“极”\end{CJK} to ``radical", which does not even appear in the reference. 
I personally think ``extremely" is a better translation.
For Example 3, the reference seems not consistency in ``how are we" and ``how do we", but our proposed system prefers to keep in consistency using ``how do we".

\begin{table}[t]
\scriptsize
\centering
%\resizebox{\columnwidth}{!}{
\begin{tabular}{c|l}
\hline
\multirow{3}{*}{Src} & \begin{CJK}{UTF8}{gbsn}养殖金枪鱼的饲料转换率是15比1。这个意思是说，每生产\end{CJK} \\
 & \begin{CJK}{UTF8}{gbsn}1磅金枪鱼肉耗费15磅用其他野生鱼类做的饲料。这可不是\end{CJK} \\
 & \begin{CJK}{UTF8}{gbsn}很具有可持续发展性。\end{CJK} \\
\hline
\multirow{2}{*}{Ref} & It's got a feed conversion ratio of 15 to one. That means it takes \\
 & fifteen pounds of wild fish to get you one pound of farm tuna. \\ 
 & Not very sustainable. \\
\hline
\multirow{3}{*}{Sys0} & Feeding tuna is 15 to one. That means that every pound of tunas \\
 & costs 15 pounds to feed feed on other wild fish. It's not \\
 & sustainable. \\
\hline
\multirow{3}{*}{Sys1} & It's 15 to 1. What that means is that every pound-pound tuna \\
 & produces 15 pounds of feed on every other wild fish. It's not \\
 & sustainable. \\
\hline
\multirow{3}{*}{\textbf{Sys2}} & And the shift rate of breeding tuna is 15 to one. That means, for \\
 & every one pound of tuna, it takes 15 pounds of feeding on other\\
 & wild fish. It's not very sustainable. \\
\hline
\hline
Src & \begin{CJK}{UTF8}{gbsn}我们爱极了革新 我们爱技术，我们爱创造 我们爱娱乐\end{CJK} \\
\hline
\multirow{2}{*}{Ref} & We \textcolor{orange}{love} innovation. We \textcolor{orange}{love} technology. We \textcolor{orange}{love} creativity. \\ 
 & We love entertainment. \\
\hline
\multirow{2}{*}{Sys0} & We love radical innovation. We love technology. We love \\
 & creation. We love entertainment. \\
\hline
\multirow{2}{*}{Sys1} & We love to be innovative. We love technology. We love to \\
 & create. We love entertainment. \\
\hline
\multirow{2}{*}{\textbf{Sys2}} & We love innovation. We love technology. We love creating.\\
 & We love entertainment. \\
\hline
\hline
\multirow{3}{*}{Src} & \begin{CJK}{UTF8}{gbsn}想要喂饱这个世界？让我们开始问：我们怎么去喂养我们\end{CJK} \\
 & \begin{CJK}{UTF8}{gbsn}自己？或者更好的，我们怎么去建立一种环境它可以让每\end{CJK} \\
 & \begin{CJK}{UTF8}{gbsn}一个团体去养活自己？\end{CJK} \\
\hline
\multirow{3}{*}{Ref} & Want to feed the world? Let’s start by asking: how are we \\
 & going to feed ourselves? Or better: how can we create \\
 & conditions that enable every community to feed itself? \\
\hline
\multirow{3}{*}{Sys0} & Do you want to feed the world? So let’s start asking: how \\
 & do we feed ourselves? Or better, how can we build an \\
 & environment that allows every group to feed themselves? \\
\hline
\multirow{3}{*}{Sys1} & How do we feed the world? So let’s start asking: how do \\
 & we feed ourselves? Or even better, how do we build an \\
 & environment that will feed itself? \\
\hline
\multirow{3}{*}{\textbf{Sys2}} & Want to feed the world? Let’s start asking: how \\
 & do we feed ourselves? Or better, how do we build an \\
 & environment that allows every single group to feed itself? \\
\hline
\end{tabular}
%}
\caption{Examples of translation results. Sys0: 1-to-1 transformer. Sys1: 3-to-3 transformer. Sys2: 3-to-3 long-short term masking transformer.}
\label{tab:examples}
\end{table}

\end{document}